\documentclass{article}

\usepackage{arxiv}

\usepackage[utf8]{inputenc} % allow utf-8 input
\usepackage[T1]{fontenc}    % use 8-bit T1 fonts
\usepackage{hyperref}       % hyperlinks
\usepackage{url}            % simple URL typesetting
\usepackage{booktabs}       % professional-quality tables
\usepackage{amsfonts}       % blackboard math symbols
\usepackage{nicefrac}       % compact symbols for 1/2, etc.
\usepackage{microtype}      % microtypography
\usepackage{lipsum}
\usepackage{graphicx}       %

\title{Face Identification Proficiency Test Designed Using Item Response Theory}

\author{
  Géraldine Jeckeln \\
  The University of Texas at Dallas\\
   \And
  Ying Hu \\
  The University of Texas at Dallas\\
     \And
 Jacqueline G. Cavazos \\
  The University of Texas at Dallas\\
     \And
 Amy N. Yates \\
  National Institute of Standards and Technology\\
  \And
   Carina A. Hahn \\
  National Institute of Standards and Technology\\
       \And
          Larry Tang \\
  University of Central Florida\\
       \And
         P. Jonathon Phillips\\
  National Institute of Standards and Technology\\
       \And
 Alice J. O'Toole \\
  The University of Texas at Dallas\\
}

\begin{document}
\maketitle

\begin{abstract}
Measures of face-identification proficiency are essential to ensure accurate and consistent performance by professional forensic face examiners and others who perform face-identification tasks in applied scenarios. Current proficiency tests rely on static sets of stimulus items, and so, cannot be administered validly to the same individual multiple times. To create a proficiency test, a large number of items of ``known'' difficulty must be assembled. Multiple tests of equal difficulty can be constructed then using subsets of items. We introduce the Triad Identity Matching (TIM) test and evaluate it using Item Response Theory (IRT). Participants view face-image ``triads'' (N=225) (two images of one identity, one image of a different identity) and select the different identity. In Experiment 1, university students (N=197) showed wide-ranging accuracy on the TIM test, and IRT modeling demonstrated that the TIM items span various difficulty levels. In Experiment 2, we used IRT-based item metrics to partition the test into subsets of specific difficulties. Simulations showed that subsets of the TIM items yielded reliable estimates of subject ability. In Experiments 3a and 3b, we found that the student-derived IRT model reliably evaluated the ability of non-student participants and that ability generalized across different test sessions. In Experiment 3c, we show that TIM test performance correlates with other common face-recognition tests.
In summary, the TIM test provides a starting point for developing a framework that is flexible and calibrated to measure proficiency across various ability levels (e.g., professionals or populations with face-processing deficits).
\end{abstract}

\section{Introduction}

\label{intro}Face identification is a critical element of law enforcement and criminal justice in the United States and abroad. This important task is carried out by professional forensic face examiners. Therefore, it is incumbent on the justice system to ensure that professional face examiners exhibit optimal performance on face-identification tasks and that their proficiency is sustained over time. However, limited work has been done to develop tests that enable the assessment of proficiency (a person's internal ability for face identification which can be inferred from their accuracy on a test) across time. To address this, we propose a novel framework that enables face-identification testing across time and across individual ability levels.
 
Reliable measures of proficiency at different points in time are desirable for several reasons. For example, individuals in professions that require them to make face identification decisions (e.g., forensic face examiners in law enforcement) will often participate in training courses designed to improve their accuracy \cite{towler2019professional}. Proficiency measures gathered over time can gauge the effectiveness of these training programs (e.g., scores before vs. after training)\cite{towler2019professional, towler2014evaluating, towler2021diagnostic}. Also, these measures can be used to assess the effects of experience and age on proficiency to assure sustained proficiency over time. To measure accuracy at different time points, we need to conceptualize a proficiency test in terms of multiple subsets of equal difficulty.

We were motivated by the special problems of forensic examiners due to their role in social justice and safety. However, to develop the framework, we studied the skills of people from the general population (e.g., undergraduates and employees from the National Institute of Standards and Technology), with the goal of developing a test that can be applied across a wide range of participant abilities. This variability would make the test applicable to people of high ability (e.g., super-recognizers, \cite{noyes2017face,ramon2019super,young2019we}), as well as to clinical populations in which individuals exhibit atypical face processing skills, for example, 
autism spectrum disorders \cite{dawson2005understanding} and schizophrenia \cite{marwick2008}.

A proficiency test should have two properties. First, it should support the creation of multiple subsets of equal difficulty. Subsets are needed, because a single test cannot be taken more than once. Repeated exposure to the same faces can inflate identification accuracy via familiarity effects \cite{roark2006learning}. This repeated exposure is highly problematic when evaluating training programs, because it can appear as if a test takers' general skills have improved, when increased accuracy is due to familiarity with the stimuli.

Previous studies 
(e.g., \cite{towler2019professional} \cite{towler2021diagnostic}) have addressed this issue by separating existing {\it face-matching tests} (e.g., Glasgow Face Matching Test [GFMT], \cite{burton2010glasgow}; Expertise in Facial Comparison Test [EFCT], \cite{white2015perceptual}) and image sets (e.g., Good, Bad and Ugly [GBU], \cite{phillips2012good})
into subsets of equal difficulty. Face matching is the most commonly used task for assessing the proficiency of professional face examiners. In these tests, individuals compare two face images (e.g., security camera image vs. mugshot) and must indicate whether the images show the same person or different people. The tests require either a binary response (``same'' or ``different'' person) (e.g., GFMT) or a response rating (e.g., -2: Sure they are the different; +2: Sure they are the same) (e.g., EFCT). In one example of subsetting existing tests, \cite{towler2019professional} measured individuals' performance on GFMT and GBU sub-tests before and after training to evaluate the effectiveness of eleven professional training courses. Similarly, in a later study, \cite{towler2021diagnostic} employed EFCT subsets to examine the effect of diagnostic feature training (i.e., training to rely on ears and facial marks) on face matching performance.

Second, a proficiency test should be calibrated. That is, it should contain stimulus items of ``known'' difficulty that can be stratified into graded difficulty levels. Consequently, subsets can be tailored to individuals of specific ability levels by sampling items (without replacement) of specific difficulty levels. This method enables the elimination of items that are too easy or to difficult for a targeted ability group. To build a calibrated test that can be separated into subsets of equal difficulty requires a large pool of items occupying a wide range of difficulty levels.

In previous applications, the difficulty of individual face-matching items were derived from human performance (e.g., proportion of test takers who have endorsed a correct response to the given item). As mentioned, common face-matching tests require identification decisions to be expressed via binary or rated response options. Although feasible, measuring item difficulty based on individuals' binary or rated face-identification decisions can be confounded by potential response bias (i.e., a user's internal tendency to select one response category over another). Bias can be due to the observer's internal decision criterion \cite{macmillan2005detection,prins2016psychophysics}, differential use of the Likert-type scale \cite{hu2017person,phillips2018face}, or to situational factors such as the perceived cost of certain types of incorrect decisions (misdentify or fail to identify). It is important to note that response bias at the level of an individual item cannot be controlled by signal detection measures, because an item is {\it either} a same-identity or different-identity item. The former can generate hits, but not false alarms; the latter can generate false alarms, but not hits.  

To illustrate how response bias complicates item difficulty measures, let's consider a face identification task with binary response options (i.e., ``same'' or ``different'' identity). When uncertain about an identification decision, an observer with a conservative response bias will exhibit a greater tendency to respond ``different identity'' in comparison to an observer with a liberal response bias. Considered from the perspective of item difficulty, a conservative response bias results in greater accuracy for different-identity pairs than for same-identity pairs. Thus, different-identity pairs would appear (incorrectly) to be easier than same-identity pairs \cite{hu2017person}. The opposite is true for liberal observers. Alternately, when a response rating is made on a Likert scale, item difficulty would be gauged by relative ``confidence'' for same- versus different-identity items. For instance, a same-identity pair that receives a response of +1 (Think they are the same) would be assumed to be more difficult than a same-identity pair that receives a response of +2 (Sure they are the same). 

Consequently, for identity-matching tasks, observer criterion and item-difficulty measures are co-dependent. This is true regardless of whether participants make binary or rated responses. This is a serious problem for cases in which groups of participants are compared. Specifically, when there are group-based differences in response bias (e.g.,  students, forensic examiners), item difficulty comparisons across groups are not valid \cite{hu2017person}.  

Previous studies do, in fact, show group-based differences in the use of response scales. Forensic examiners, compared to untrained undergraduates, concentrate their responses in the middle of the scale (less certain), thereby avoiding ``high confidence'' responses at extreme ends of the scale \cite{hu2017person,phillips2018face}. Examiners may adopt this strategy to avoid the repercussions of high-confidence misidentifications in forensic face settings. This compromises the validity of item difficulty measures applied across groups of individuals (e.g., forensic examiners, students). Again, from the perspective of item difficulty, many items would be found to be more difficult for examiners (higher-ability individuals) than for students (lower-ability individuals) \cite{hu2017person,white2015perceptual}, when instead the item difficulty measure is driven by the differential use of the scale by the two populations. One theoretical approach to measuring item difficulty directly is to use Item Response Theory (IRT, \cite{lord1980applications}). Before presenting the main part of the study, we first introduce IRT for assessing item difficulty and participant ability. 

%\subsection{Item Response Theory}

In recent research on face perception \cite{wilmer2012capturing,cho2015item,sunday2018age,thomas2018signal}, IRT has been proposed as a method for test evaluation and ability assessment. IRT is a psychometric theory used to model the association between face-identification decisions (participant responses to items) and face-identification ability.

IRT encompasses a group of latent variable models that link item responses (e.g., face-identification judgment) to a single latent variable (e.g., face-identification ability) \cite{rizopoulos2006ltm}. In the case of dichotomous items (the response can be correct or incorrect), IRT models are used to compute the probability of a correct response endorsed by the $i^{th}$ participant on the $j^{th}$ item. An IRT model is fit to item responses and is expressed as follows: 

\begin{equation}\label{generalmodel}
 P(x_{ij}=1 | \theta_i) = c_j + (1- c_j)g\{\alpha_j,(\theta_i- \beta_j) \},
 \end{equation}

\noindent
    where $x_{ij}$ represents the response status (1= correct, 0= incorrect) of the $i^{th}$ participant on the $j^{th}$ item, $\theta_i$ denotes the participant latent score (e.g., ability), $\emph{c}_j$ denotes the item guessing parameter,  $\alpha_j$ denotes the item discrimination parameter, and $\beta_j$ denotes the item difficulty parameter. The slope of the line ($\alpha_j$) determines the item's sensitivity to changes across the latent scale (the steeper, the better at discriminating participants of a different ability  levels). Item difficulty (intercept)  ($\theta_j$) determines the location on the latent scale that yields a 0.5 probability of correct response. The lower asymptote of the line ($\emph{c}_j$) is used to represent a correct answer endorsed by guessing. In this paper, we considered the one-parameter logistic model (Rasch model \cite{wright1977solving}), where $\alpha_j$ is constrained to a value of 1, and $\emph{c}_j$ is constrained to a value of 0. However, additional models are available for estimating these parameters for dichotomous items \cite{rizopoulos2006ltm}. The two-parameter logisitic model estimates both $\beta_j$ and $\alpha_j$, while keeping $\emph{c}_j$ constrained to a value of 0. The three-parameter logistic model computes estimates for all three item parameters ($\alpha_j$, $\beta_j$, and $\emph{c}_j$).

\begin{figure}[ht!]
\begin{center}
\includegraphics[width=\textwidth]
{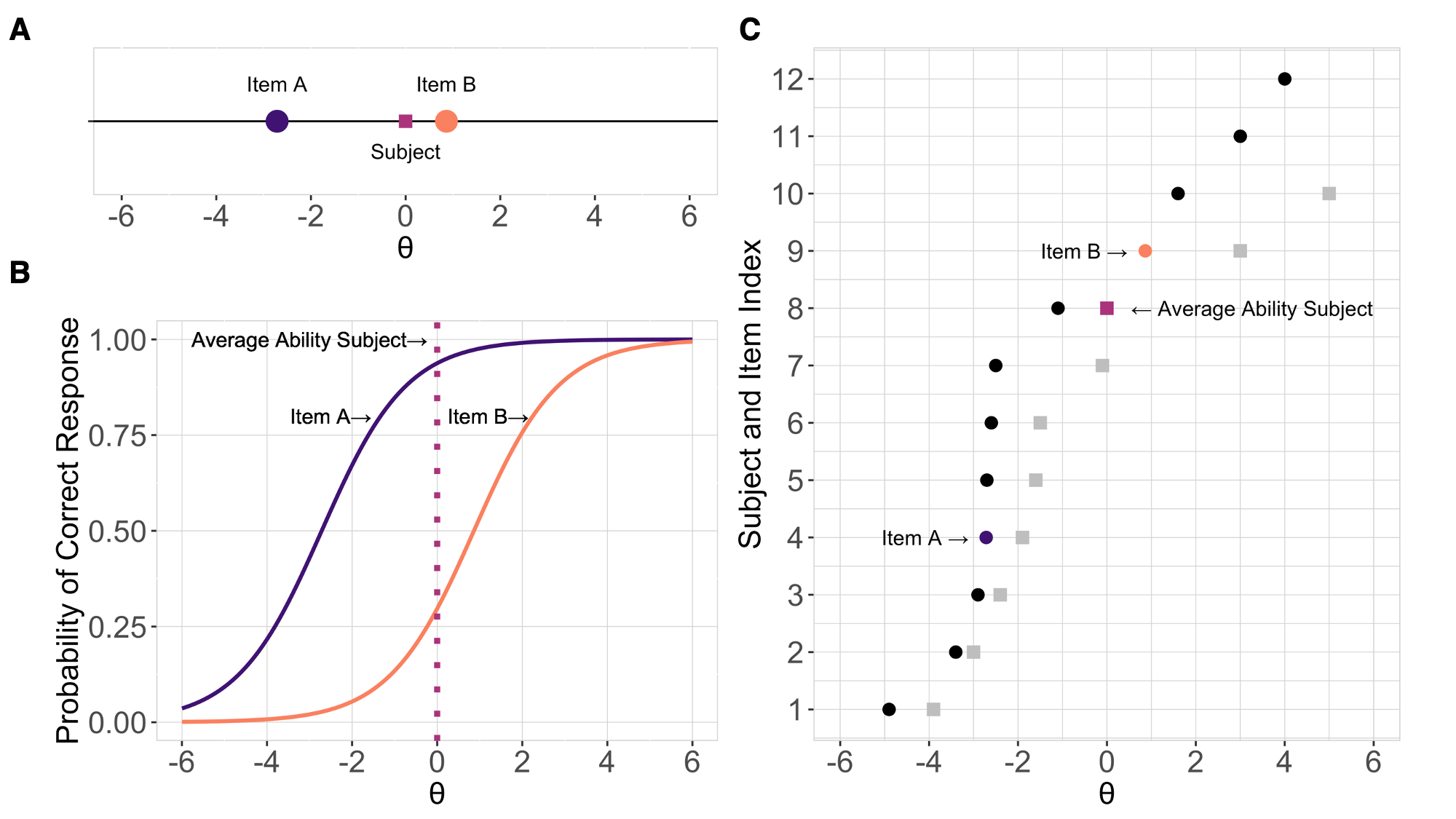}
\caption{A) Example of IRT Subject and Item Scale. Item difficulty ($\beta$) and subject ability ($\theta$) occupy the same latent scale ranging from low (easy item, poor ability) to high (difficult item, high ability). By convention, this shared scale is labeled ($\theta$). In the model used in this paper, average ability is defined as 0. One exemplar subject (magenta square) is used to represent average ability. If a subject’s estimated ability (square) is greater than the estimated difficulty of an item (circles), the subject has an above-chance probability to answer the given item correctly. This can be seen more clearly when the probability of a correct response is plotted as the function of subject ability. B) For example, a subject with average ability (dotted line, $\theta$=0) has an above-chance probability (.94) of endorsing a correct response to Item A (B, $\beta$ = -2.72) and below-chance probability (.30) of endorsing a correct response to Item B (C, $\beta$ = 0.86). C) Exemplar items (circles) and subjects (squares) plotted along the ability and difficulty scale. The index consists of items and subjects ranked by difficulty and ability, respectively.}
\label{shared_scale}
\end{center}
\end{figure}

IRT offers a promising route for testing identification ability, because it provides estimates of ability based on the properties of items. Also, IRT has several features that are critical for developing a face-identification test. We consider these in turn. First, IRT provides measures of participant ability and item difficulty that occupy the same scale \footnote{Following conventions used in the IRT literature, we refer to this as the ``Theta scale,'' henceforth labeled $\theta$.} and can be compared directly to one another \cite{de2013theory}. This property is particularly valuable for building assessment tools that are intended to capture specific levels of face-identification ability. As illustrated in Fig.~\ref{shared_scale}, this important feature enables the user to infer each participants' probability of responding to a specific item correctly, given their respective position on the ability scale. 
 
Second, IRT provides item-difficulty measures that are independent of the participant sample \cite{de2013theory}. Concomitantly, IRT provides ability measures that are independent of the item sample and can be generalized to the participants' true skill level \cite{de2013theory}. 
 
Third, IRT provides precision measurements at the individual participant ability level. This feature of IRT enables evaluation of the test for assessing people of different ability via the Test Information Function \cite{de2013theory,wilmer2012capturing}. The peak of this function corresponds to the participant ability level that is best suited for evaluation with the test. For example, IRT can assess the efficiency of existing assessment tools for diagnosing individuals with impaired face recognition \cite{cho2015item}.

%comment out for now 

In previous studies, IRT was used to evaluate the quality of face-recognition tests \cite{cho2015item}, to isolate face-recognition ability from other abilities \cite{wilmer2012capturing}, and to assess item bias towards certain demographic groups \cite{sunday2018age}. These previous studies assess \textit{memory} for faces. IRT has not yet been applied to face-identification tasks that are based on \textit{perception} (without memory requirements) (See Supplemental Material, Figure 1). These perceptual abilities are tapped in forensic identification (e.g., identity matching). Here, we use IRT to analyze the results of face-identification tests that rely on perceptual abilities.

Specifically, our goal for this study was to develop a face-identification test that enables testing across time and across ability levels. To achieve this goal, first we propose a three alternative forced choice (3-AFC) face-identification test (the Triad Identity Matching [TIM] test). A 3-AFC paradigm can be used to construct calibrated subsets of items, because it allows for item difficulty estimates that circumvent the response bias issues of existing tests. Therefore, the second step was to use IRT to measure the psychometric properties of the TIM test. Item-difficulty scores extracted from the IRT modelling were used to create subsets of stimulus items that can be partitioned into equal difficulty levels or stratified into various graded difficulty levels (“easy” or “difficult”). To date, no study has considered the usefulness of IRT for evaluating the psychometric properties of face-matching tasks. 

The following text is organized into the following sections. In Section 2, we describe the TIM test construction. In Section 3 (Experiment 1), we provide data from university students on the TIM test and evaluate the test using IRT, along with traditional measures of item difficulty and subject accuracy (proportion correct). In Section 4 (Experiment 2), we demonstrate how IRT can be used to guide the construction of equally difficult subsets, and provide comparisons between ability estimates computed from the subsets of items and the full test.
In Section 5 (Experiment 3), we examine the generalizability of the TIM test across a different group of participants, across separate experimental sessions, and across different commonly used face-recognition test. In Section 6 (Discussion), we conclude with the contributions and limitations of this work.

\section{Section 2: TIM Test Construction}
\label{sec:Methodology}

\begin{figure}[ht!]
\begin{center}
\includegraphics[width=\textwidth]{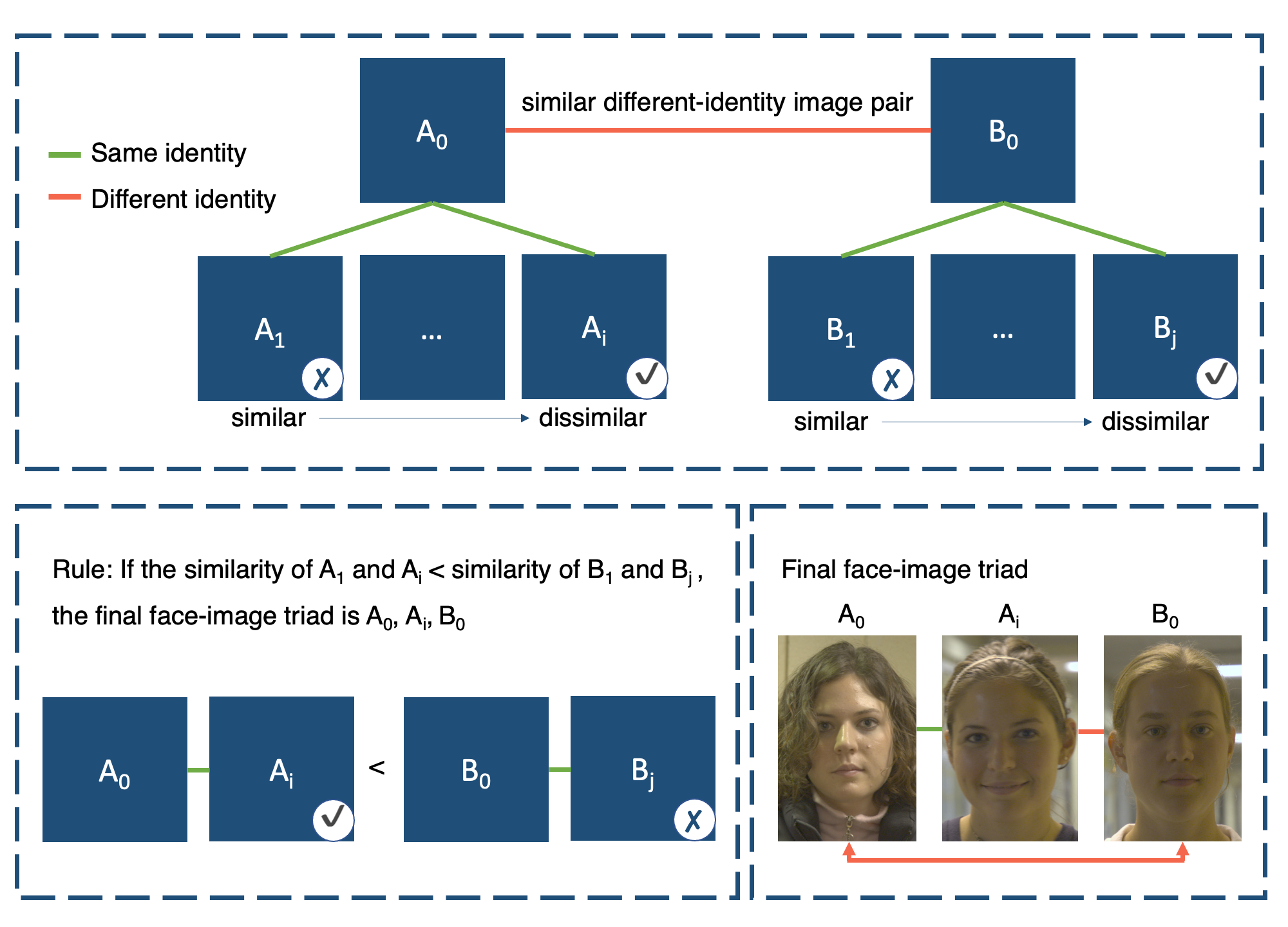}
\caption{Stimuli screening paradigm. Image pairs were divided into same-identity and different-identity pairs. Then, different-identity pairs were ranked from similar to dissimilar using similarity scores obtained from a deep convolutional neural network (VGG-face, \cite{parkhi2015deep}). All different-identity pairs were demographically constrained (``yoked'') so that only the same-race and same-gender pairs remained. Next, for each identity, only the different-identity pair with the largest similarity score was chosen. For each identity within a different-identity pair, a same-identity image with the lowest similarity was selected. The identity with the lowest same-identity pair similarity score was selected to be part of the triad. Therefore, each triad consisted of the lowest different-identity pair and the highest same-identity pair for that identity.}
\label{stimuli_screening_paradigm}
\end{center}
\end{figure}

We created the TIM test, a 3-AFC test, consisting of image triads: two same-identity images and one different-identity image. Participants determine which of the images depicts the different-identity (``odd-one-out'') (Figure \ref{stimuli_screening_paradigm}). A total of 225 triads were created using 675 images sampled from the {\it Good, Bad, and Ugly Face Challenge Dataset} \cite{phillips2012good}. Images were taken in frontal view and varied in illumination, expression, and participant appearance (e.g., accessories and hair).

To avoid ceiling effects, triads were constructed to minimize the similarity of images that showed the same identities. The different-identity image in the triad was chosen to be as similar as possible to one of the same-identity images 
(Figure \ref{stimuli_screening_paradigm}). Images for triads were selected using data from VGG-Face \cite{parkhi2015deep}. We selected this algorithm, because it proved comparable in ability to students in a face identity matching test (cf. A2015 [\cite{phillips2018face}). Here, we used the top-level face descriptors from the algorithm to compute similarity scores between images. In what follows, for any identity pair A and B, a triad includes two images of one identity ($A_0$ and $A_i$) and one image of a different identity ($B_0$).

Figure \ref{stimuli_screening_paradigm} shows a more detailed account of the following step-wise process for triad construction.

\begin{enumerate}
    \item Image pairs were divided into same- and different-identity pairs. 
    \item Similarity scores were used to rank pairs of different-identity images from the most to the least similar. 
    \item  The different-identity image pair 
    ($A_i$, $B_j$) with the largest similarity score was selected. 
    \item  Similarity scores were also used to rank same-identity images ($A_i$, $A_j$) from the most to the least similar. The least similar image was chosen to complete the triad. Each identity appeared in 2 to 35 face images sampled from the GBU dataset (average image per identity = 15.47).
\end{enumerate}
Therefore, each triad consisted of different-identity pairs that are similar and a same-identity pair that is dissimilar ($A_0$, $A_i$, and $B_0$). By design, the algorithm should perform poorly on the triads. We verified the difficulty of the TIM test for the algorithm, as follows. We treated VGG-face as a participant and simulated the 3-AFC face identification task. The odd-one-out decision was made using the algorithm-generated similarity scores. For each triad, the two images with the highest similarity were judged as the same identity and the remaining image was selected as the odd-one-out. The selection was compared to the ground truth and the proportion correct was calculated. The result showed systematically incorrect performance (proportion correct = .1378) for the algorithm, thereby supporting the use of similarity scores from VGG-face \cite{parkhi2015deep} to construct highly challenging triads. 

\section{Section3: Experiment 1 - Normative Performance on the TIM Test}
\label{sec:Experiment 1}
In Experiment 1, we demonstrate that the TIM test can capture a large range of student performance and individual item accuracy. In addition, we show that using IRT-based parameters we can obtain item difficulty and participant ability measures on the TIM test. First, student performance was evaluated on the full set of TIM stimuli. Individual student baseline accuracy was calculated as the proportion of correct responses. Second, we employed IRT modeling (Rasch, 1960, \cite{wright1977solving}) to evaluate the psychometric properties of the test. 

\subsection{Methods}
\textbf{Participants.} A total of 203 undergraduate students from The University of Texas at Dallas (UTD) participated in this study. Data collection took place during the Spring 2019 (77 participants) and (early) Spring 2020 (126 participants) semesters. Participants were recruited through The School of Behavioral and Brain Sciences online sign-up system and were compensated with research exposure credits. Participants were required to be at least 18 years of age and have normal- or corrected-to-normal vision. Two participants were excluded due to software error (data
collection impediment) and four participants were excluded due to missing data (overwritten data files). The final data included 197 participants (140 female, 55 male, and 2 indicated ``other''), ranging from age 18 to 36 (average age = 20.23). All aspects of the study were in accordance with the UTD Institutional Review Board protocol.

\textbf{Procedure}.
For each participant, data collection took place in a single experimental session and included the full TIM test (225 items), followed by a demographic survey (via Qualtrics \cite{qualtrics2013qualtrics}\footnote{Certain commercial equipment, instruments, or materials are identified in this paper to foster understanding. Such identification does not imply recommendation or endorsement by the National Institute of Standards and Technology, nor does it imply that the materials or equipment identified are necessarily the best available for the purpose.}). The experiment was programmed using PsychoPy v1.84.2 \cite{peirce2007psychopy}. For each trial, a triad was presented for 3.5 seconds. Response time was not limited and no feedback was provided. Trial order and image position within a triad were randomized across participants.

\subsection{Analysis and Results}
\textbf{Baseline Performance}. Performance was measured as the proportion of items answered correctly. Chance performance was $0.33$. Participant accuracy was well above chance (\textit{M}= 0.69, \textit{SD}= 0.11, \textit{Mdn}= .70) and ranged from 0.37 to 0.89. Item accuracy (proportion of participants who answered each item correctly) varied widely, ranging from .17 to .97 (\emph{M}= .69, \emph{SD}= .17, \emph{Mdn}= .72).

\begin{figure}[ht!]
\begin{center}
\includegraphics[width=\textwidth]
{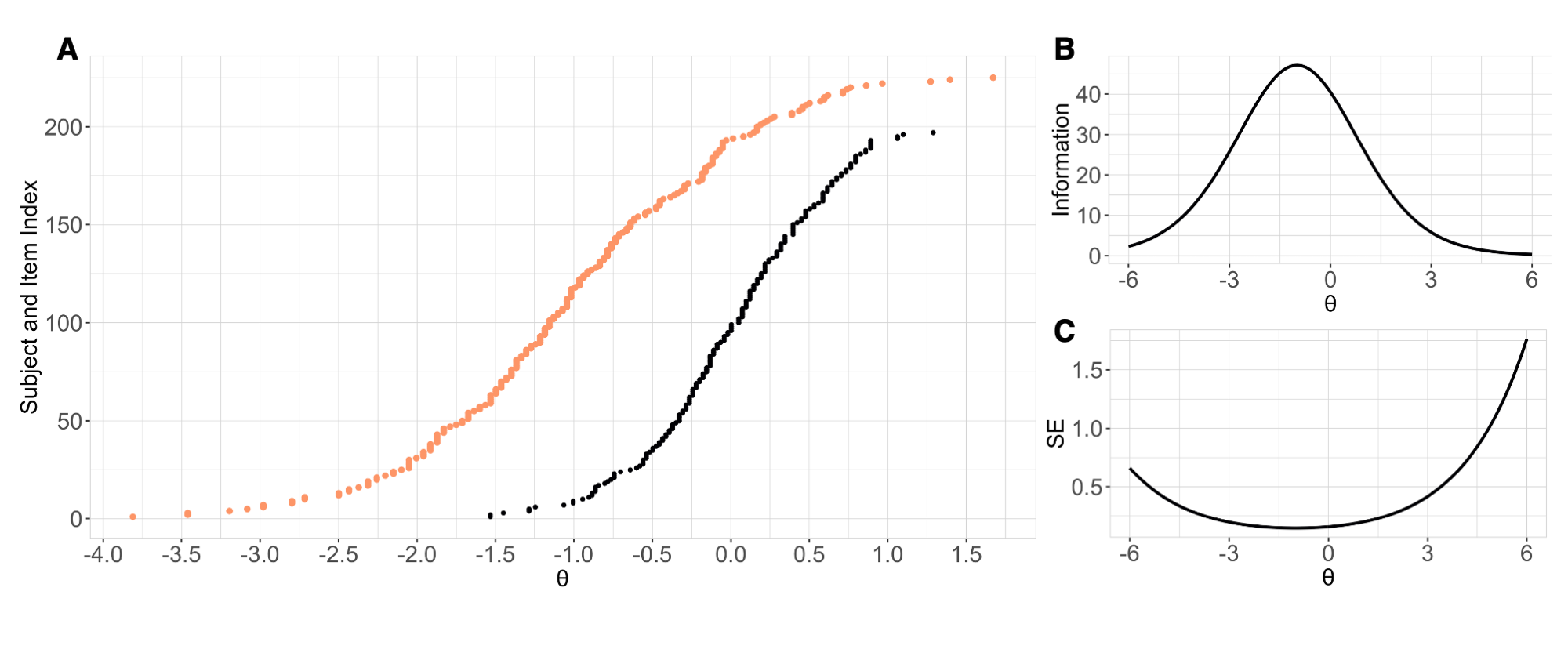}
\caption{(A) One-parameter logistic model fit to 225 items and 197 participants. Item difficulty (orange) and participant ability (black) estimates are plotted on the same scale. (B) The test information function is the reciprocal of the standard error of the estimated construct ($\theta$) and is commonly used to indicate the degree of measurement precision for any ability score. The results suggests that the TIM test is most informative for ability scores ranging between low to average. (C) The standard error of the estimated construct ($\theta$).}
\label{fig:Scree_Dots_Info_SE}
\end{center}
\end{figure}

\textbf{IRT Modeling}. We employed IRT modeling to evaluate the psychometric properties of the TIM test (see Figure~\ref{fig:Scree_Dots_Info_SE}). A one-parameter logistic model (Rasch model, \cite{wright1977solving}) was fit to the data employing Expectation Maximization (EM). All aspects of IRT modeling were conducted in R, using the mirt package v1.29 \cite{chalmers2012mirt}. A scree test \cite{beaton2014exposition} was used to evaluate the dimensionality of the data and to ensure that the TIM test measured a single latent variable (face identification ability). Model fit was assessed using the root mean square error of approximation (RMSEA), Akaike information criterion (AIC), and Bayesian information criterion (BIC). A RMSEA of .6 and below is considered a good model fit. 

Scree test results indicated unidimensional data. Results also indicated a good fit for the one-parameter logistic model (RMSEA = 0, AIC = 47195.18, BIC = 47937.19). Figure~\ref{fig:Scree_Dots_Info_SE}A shows fitting responses of 197 participants on 225 triads. Triad difficulty spanned $-3.81$ to $1.67$, and participant ability spanned $-1.53$ to $1.29$. An efficient proficiency test should capture accurate estimates of proficiency for different ability groups. To simulate this, we show that IRT models built from groups with high (low) ability individuals can accurately measure the proficiency of groups with low (high) regardless of differences in abilities (See Supplemental Materials, IRT Modeling Generalizability and Figure 2).

%For both item difficulty and participant ability, linear regions overlapped. 
The Test Information Function (TIF) Curve illustrates how well the TIM test can evaluate participant ability across different levels of ability. The peak of the TIF Curve (Figure~\ref{fig:Scree_Dots_Info_SE}B) indicates that the amount of information (I) provided by the test reaches a maximum (\textit{I}=47.11) at $\theta$ = -0.99. Note that I has a maximum at an ability score slightly below average ($\theta$ = 0). The TIF is a standard measure in IRT. However, the amount of precision with which the test measures different levels of ability can also be conveyed using more common psychological measures such as standard error (SE) of the estimated construct (Figure~\ref{fig:Scree_Dots_Info_SE}C). The SE curve is the inverse of the TIF. In agreement with the TIF curve, results indicate that the test provides participant-ability estimates with greatest precision (lowest SE) at $\theta$ = -0.99. \footnote{Baseline performance on an initially smaller sample size (N = 76), (\emph{M}= 0.68, \emph{SD}= 0.11, \emph{Mdn}= 0.69), was comparable to our final sample size (N = 197). Item difficulty measures also were correlated across participant sample sizes [proportion correct: \textit{r} = .97, \textit{p} \textless .0001; $\beta$: \textit{r} = .97, \textit{p} \textless .0001].}

\subsection{Experiment 1 Discussion}

The TIM test captured a large range of student performance and individual item accuracy. Moreover, the TIM item distribution occupies a range of difficulty that exceeds the lowest and highest ability scores exhibited by our sample of university students. These results suggest that TIM test offers a range of item difficulty that is large enough to prevent ceiling or floor effects in individuals from the general population. Additionally, the items located at each extreme of the difficulty distribution may be useful to test individuals with ability levels below and above that of our present sample. Also, the test was particularly informative for participants with ability slightly below average.

\section{Section 4: Experiment 2 - Creating Subsets of Customized Difficulty}
\label{sec: Experiment 2}
In Experiments 2a and 2b, we show that equally challenging TIM subsets can be used to estimate participant ability as effectively as using the full 225-item test. Equally-difficult subsets are crucial for recruitment and training purposes, particularly in applied scenarios (forensic facial examination). Furthermore, item subsets of graded difficulty are needed for testing participant groups of different ability. In Experiments 2a and 2b we examined whether TIM subsets produce estimates of participant performance (proportion correct and ability score) that are consistent with those derived from the full TIM test. This was tested in Experiment 2a by using subsets designed to target different ranges of participant ability (three ``Easy'' subsets for lower-ability individuals and three ``Difficult'' subsets for higher-ability individuals), and in Experiment 2b by using subsets occupying the full range of item difficulty.

\subsection{General Methods for Experiments 2a and 2b}
\textbf{Human data and IRT modelling.}
All analyses were carried out using the university student data collected in Experiment 1 and the one-parameter logistic model trained in Experiment 1. 

\textbf{Creating subsets.} For Experiment 2a, the TIM test items ($n=225$) were partitioned into six 36-item subsets as follows: First, items were ranked from most easy to most difficult based on item-difficulty measures derived from the one-parameter logistic model. Second, the ranked items were median split into an easy and a difficult set. Finally, items from each easy and difficult set were sampled randomly (without replacement) to create three ``Easy'' subsets (E1, E2, and E3) and three ``Difficult'' subsets (D1, D2, and D3). For Experiment 2b, the TIM test items ($n=225$) were partitioned into a total of three 72-item subsets of average difficulty (S1, S2, and S3). Each subset was created by combining one ``Easy'' and one ``Difficult'' subset (i.e., E1 and D1; E2 and D2, E3 and D3). Descriptive statistics for subset difficulty are reported in Table \ref{table:items}.

\begin{table}[!]
%\begin{center}
\centering
\caption{Descriptive statistics for item difficulty ($\beta$) for the subsets in Experiments 2a and 2b.}
\begin{tabular}{cccc}
\toprule
Subset & \emph{M} & \emph{SD} & \emph{Mdn} \\
\hline
\midrule
Experiment 2a\\
Easy 1 & -1.72 & .47 & -1.66 \\
Easy 2 & -1.72 & .67 & -1.48  \\
Easy 3 & -1.84 & .64 & -1.67 \\
Difficult 1 & -.30 &  .55 & -.30 \\
Difficult 2 & -.23 & .60 & .41 \\
Difficult 3 & -.18 & .55  & -.17 \\
\hline
Experiment 2b \\
Set 1 & -1.01 &  0.87 & -1.03\\
Set 2 & -0.97 & 0.98 & -1.00 \\
Set 3 & -1.01 & 1.03 & -1.03 \\
\hline
\bottomrule
\end{tabular}
\label{table:items}
%\end{center}
\end{table}

\subsection{Results for Experiments 2a}

\begin{figure}[ht!]
\begin{center}
\includegraphics
[width=0.8\textwidth]
{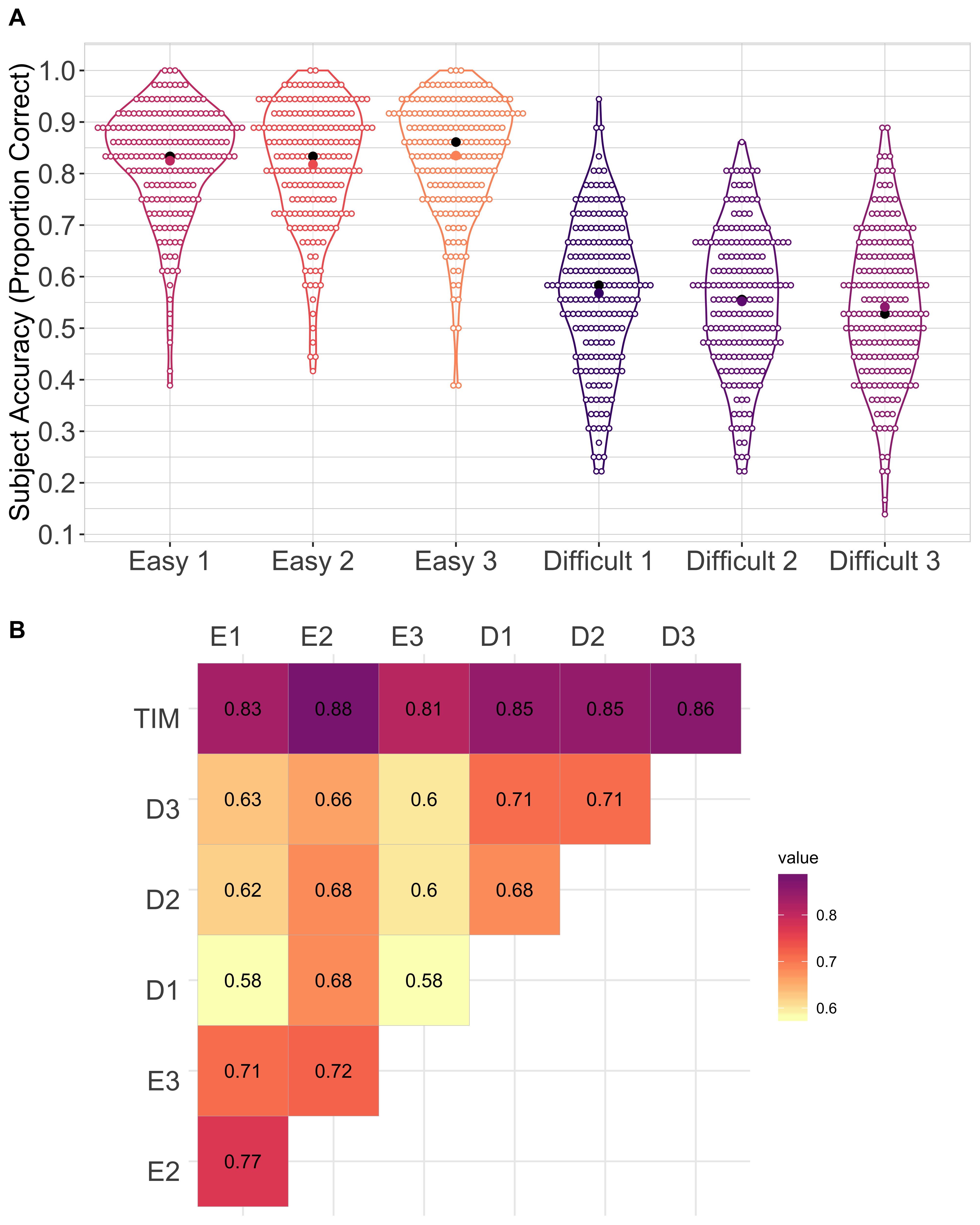} 
\caption{A) Violin plots of participant accuracy (proportion correct) on each item subset. The empty circles represent the accuracy of individual participants on each item subset, the coloured dots represent the item subset mean, the black dots represent the item subset median. B) Pearson correlation between participant accuracy (proportion correct) on the full TIM test and all subsets. The Full TIM test was highly correlated with all six item subsets. All comparisons are significant at the 0.01 level.}
\label{fig:Subsets_PC_COR}
\end{center}
\end{figure}

\textbf{Baseline accuracy.} Here, we demonstrate that human accuracy is consistent across subsets of equal difficulty, and that performance on both ``Easy'' and ``Difficult'' sets is indicative of performance on the full TIM test. Baseline performance was measured as the proportion of items answered correctly for each subset. Descriptive statistics of human performance are reported in Table \ref{table:participants} and plotted as a violin plot in Figure \ref{fig:Subsets_PC_COR}A. As expected, accuracy was higher for the ``Easy'' subsets than for the ``Difficult'' subsets. In addition, we compared proportion correct on the subsets against proportion correct on the full test. Pearson product-moment correlation results indicated a strong positive relationship between the full TIM item bank and each subset of items (\textit{r} = .81 - .88)(see Figure \ref{fig:Subsets_PC_COR}B). Comparisons across ``Easy'' subsets showed a moderate positive relationship, ranging from (\textit{r} = .71) to  (\textit{r} = .77). Comparisons across ``Difficult'' subsets showed a moderate positive relationship, ranging from (\textit{r} = .68) to  (\textit{r} = .71).

\begin{table}[htpb]
\centering
\caption{Descriptive statistics for participant accuracy (proportion correct) on the subsets in Experiments 2a and 2b.}
\begin{tabular}{cccc}
\toprule
Subset & \emph{M} & \emph{SD} & \emph{Mdn} \\
\hline
\midrule
Experiment 2a \\
Easy 1 & .83 & .11 &.83 \\
Easy 2 & .82 & .12 & .83 \\
Easy 3 & .84 & .11 & .86 \\
Difficult 1 & 57 & .14 & .58\\
Difficult 2 & .55 & .14 & .56 \\
Difficult 3 & .54 & .14 & .53\\
\hline
Experiment 2b \\
Set 1 & 0.70 & .11 & .71 \\
Set 2 & .68 & .12  & .69 \\
Set 3 & .69 & .11 & .69\\

\hline
\bottomrule
\end{tabular}
\label{table:participants}
\bigskip
\end{table}

\begin{figure}[ht!]
\begin{center}
\includegraphics[width=\textwidth]
  {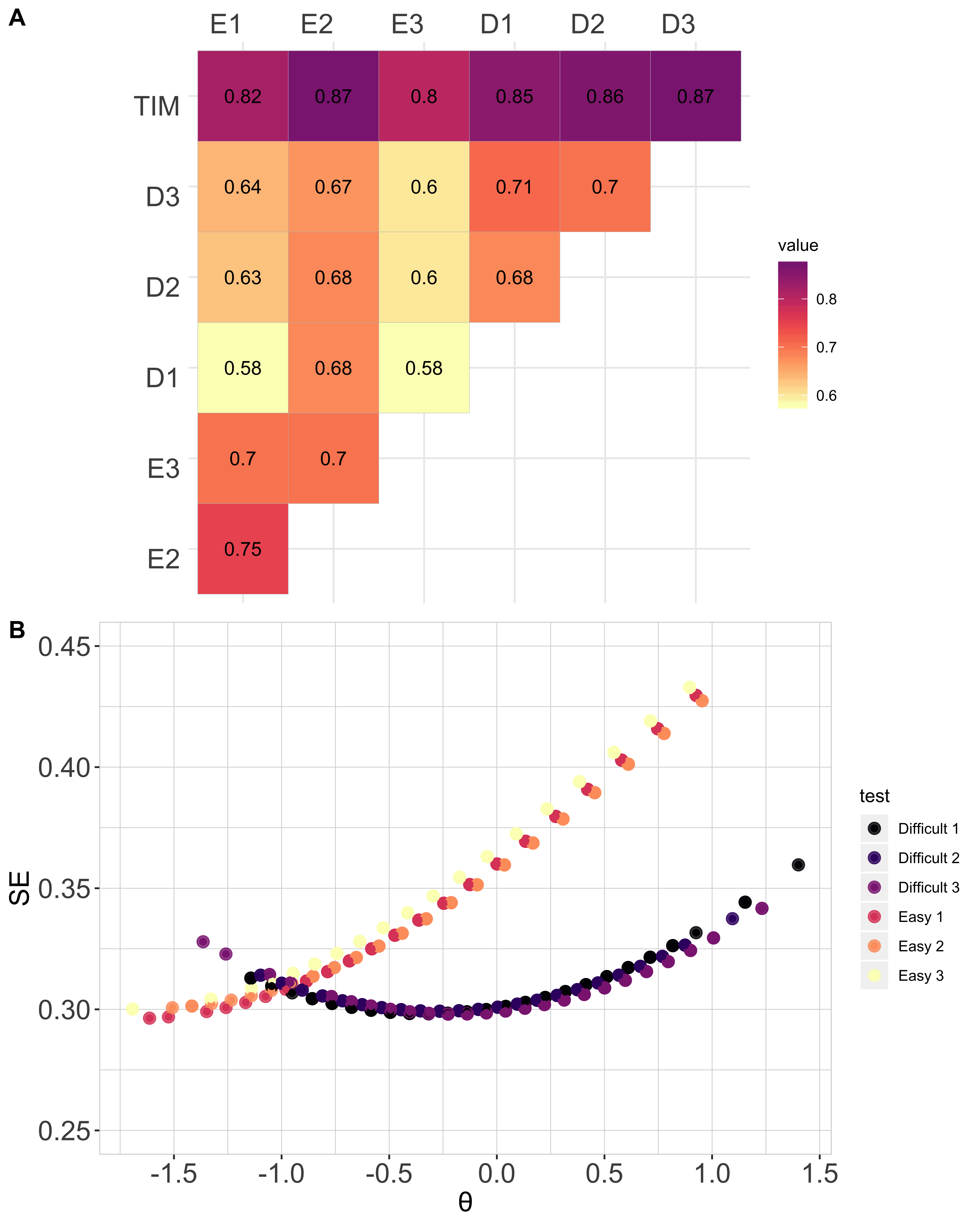}
\caption{A) Pearson correlation between participant ability on the full TIM test and all subsets. All comparisons are significant at the 0.01 level. B) Standard error of the ability estimate for all participants on all subsets.}
\label{fig:Subsets_Theta_COR_SE}
\end{center}
\end{figure}

\textbf{IRT-based estimates of ability.} Here, we demonstrate that the TIM test and the one-parameter logistic model trained in Experiment 1 produce consistent estimates of participant ability with different test sizes. Specifically, we show that smaller subsets of TIM items converge to give similar estimates of participant ability. This analysis was carried out as follows. First, for a given item subset (e.g., Easy 1), we retrieved the responses for all participants in Experiment 1 ($n = 197$). Next, the responses (e.g., Easy 1: 197 participants x 36 items) were projected to the model trained in Experiment 1 (full set of TIM items: 197 participants x 225 items). This resulted in a new set of ability scores, estimated by the model trained on the full set of items and using responses to a selected set of items (e.g., Easy 1). These steps were repeated for each item subset. Finally, the ability scores estimated from each item subset were compared to the ability scores estimated using the full TIM test (Experiment 1). Results indicated strong positive correlations between participant ability estimated from the full test and participant ability estimated from the subsets, which  ranged from (\textit{r} = .58) to (\textit{r} =.87) (see Figure \ref{fig:Subsets_Theta_COR_SE}).  This range of results is as expected and is consistent for an IRT model that fits well.

To evaluate the level of precision with which the ``Difficult'' and ``Easy'' subsets estimate participant ability, we plotted standard error of the ability estimate for each participant on each subset in Figure \ref{fig:Subsets_Theta_COR_SE}B. Overall, standard error estimates were lower for all three difficult subsets, which suggests that these three difficult subsets provide more reliable measures of ability.

%comment out for now 

\subsection{Results for Experiments 2b}

\begin{figure}[ht!]
\begin{center}
\includegraphics
[width=0.8\textwidth]
{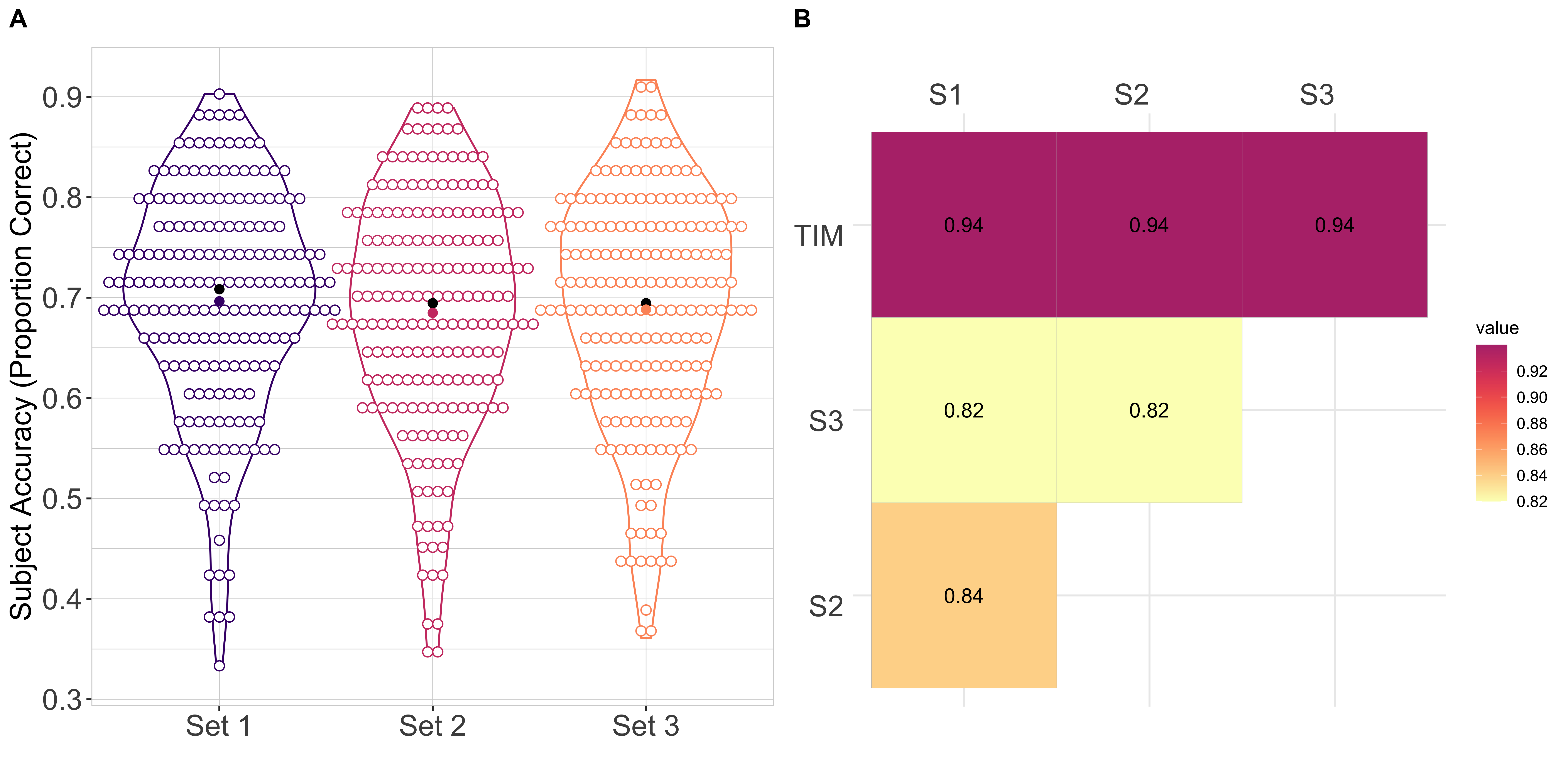} 
\caption{A) Violin plots of participant accuracy (proportion correct) on each item subset. The empty circles represent the accuracy of individual participants on each item subset, the coloured dots represent the item subset mean, the black dots represent the item subset median. B) Pearson correlation between participant accuracy (proportion correct) on the full TIM test and all subsets. 
}
\label{fig:Subsets_PC_COR_2b}
\end{center}
\end{figure}

\textbf{Baseline accuracy.} In Experiment 2b, we repeated the analyses reported in Experiment 2a using Sets 1, 2, and 3. As expected, participant performance (proportion correct) was comparable across the three sets and the TIM test (See Table \ref{table:participants}). We compared proportion correct on the three sets against proportion correct on the full test. Pearson product-moment correlation results indicated a strong positive relationship between the full TIM item bank and each subset of items (\textit{r} = .94)(see Figure \ref{fig:Subsets_PC_COR_2b}A). Comparisons across all sets showed a strong positive relationship, ranging from (\textit{r} = .82) to  (\textit{r} = .84). This range of results is as expected and is consistent for an IRT model that fits well.

\begin{figure}[ht!]
\begin{center}
\includegraphics[width=\textwidth]
  {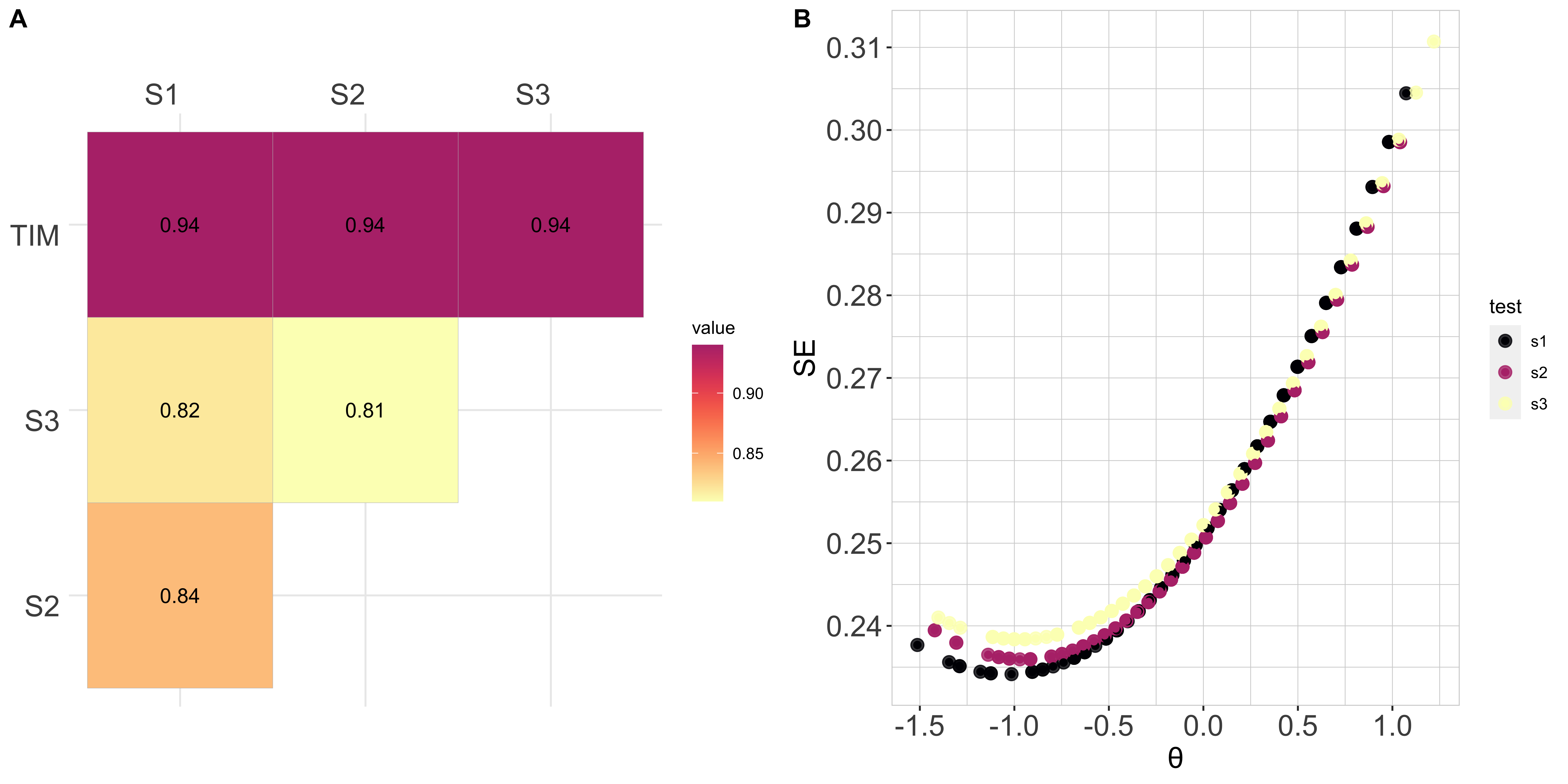}
\caption{A) Pearson correlation between participant ability on the full TIM test and all subsets. All comparisons are significant at the 0.01 level. B) Standard error of the ability estimate for all participants on all subsets.
}
\label{fig:Subsets_Theta_COR_SE_2b}
\end{center}
\end{figure}

\textbf{IRT-based estimates of ability.} 
We repeated the analyses reported in Experiment 2a using Set 1, Set 2, and Set 3. As expected, the results showed a strong positive relationship between participant ability estimated from the full test and participant ability estimated from the subsets (\textit{r} = .94). Additionally, the results showed a strong positive relationship between participant ability estimated across subset ranging from (\textit{r} = .81) to (\textit{r} =.84) (see Figure \ref{fig:Subsets_Theta_COR_SE_2b}A). Standard error estimates were comparable across subset (See Figure \ref{fig:Subsets_Theta_COR_SE_2b}B). Consistent with the results pertaining to the full test (Experiment 1), the TIM subsets provide measures of proficiency with the highest level of precision for participants with ability slightly below average (See Figure \ref{fig:Subsets_Theta_COR_SE_2b}B)

\subsection{Experiment 2 Discussion}
In Experiment 2, we provide a proof of principle of the validity of sub-sampling TIM items for evaluating individual proficiency. In Experiment 2a, we created six 36-item subsets of specific challenge levels (three ``Easy'' subsets and three ``Difficult'' subsets) intended to target different ranges of participant ability. The subsets provided measures of proficiency that were consistent with measures derived from the entire TIM test (225-item set). In addition, we demonstrate the usability of IRT for providing precision measurements for individual ability estimates. Using this feature we show that  the ``Difficult'' subsets yielded measures of individual ability that were more reliable (smaller error rate) than those estimated using the ``Easy'' subsets. The results indicate also that all subsets yielded the most reliable ability estimates for participants ranging between low to average ability.
In Experiment 2b, we repeated these analyses using three 72-item sets of average difficulty. Consistent with Experiment 2a, participant-ability estimates were consistent across test size (72-item sets and full TIM test). Moreover, the three subsets provided precision measures comparable to using the entire TIM test.
Together, these findings suggest that the 225-item TIM test is a viable tool for creating subsets, with known difficulty and precision properties, aimed at evaluating individual performance across different points in time. This methodology can be used for evaluating ability increases that result from training programs, by administering subsets of equal difficulty before and after training sessions. This methodology can be used also for assessing stability in ability across time in the absence of training. Further research is required to test the usability of specific challenge level for specific trainee groups.

\section{Section 5: Experiment 3 - Generalizability of the TIM test}

In Experiment 3, we evaluated the generalizability of the TIM test. We examined generalizability in terms of participant population (Experixment 3a), generalizability in performance across testing session (Experiment 3b), and comparability with established face-matching and face-memory tests (Experiment 3c). We begin by demonstrating that the TIM test results remain consistent across a different population of human observers (federal employees versus university students) and different experimental setting (National Institute of Standards and Technology [NIST] versus UTD). In what follows, we show that the test occupied a large range of human and item accuracy, and that non-student participants can be evaluated on TIM subsets using an IRT model trained on a more common and accessible participant sample (university students) and a larger set of items (full TIM test) (Experiment 3a). Next, we evaluated participants across two separate testing sessions (using two equally-difficult tests) and demonstrate that individual performance varies less across testing sessions than across tests (Experiment 3b). Lastly, we demonstrate that human ability estimated using the TIM test is correlated with human performance on commonly used tests of face-matching and face-memory ability (Experiment 3c).

\subsection{General Methods for Experiment 3a, 3b, and 3c}
Experiments 3a, 3b, and 3c, were conducted across two separate testing sessions separated by approximately one week. Each testing session included a selection of face-recognition ability tests. We begin by introducing the general methods employed across all three experiments.

\textbf{Participants.} A total of 58 federal employees from the NIST participated in Experiment 2. Data were collected from August 2019 to March 2020. Participants were recruited verbally and via flyers posted throughout the NIST Gaithersburg campus. Two participants were removed from the final analysis due to computer error. The final sample included 56 participants (30 female, 26 male). The majority of participant self-identified as White ($n = 42$), other participants identified as Asian ($n = 7$), Black or African American ($n = 5$), Native Hawaiian or Other Pacific Islander ($n = 1$) and mixed-race ($n = 1$). Four participants identified as Hispanic or Latino and 52 did not identify as Hispanic or Latino. Age composition of the participant sample can be found in Table \ref{table:participant_demographic}.

\begin{table}[htpb]
\centering
\caption{Participant demographic (age composition)}
\begin{tabular}{cccc}
\toprule
Age group & \emph{N}  \\
\hline
\midrule
18 to 29 &	6 \\
30 to 39 &	15 \\
40 to 49 &	9 \\
50 to 59 &	20 \\
60 plus &	6 \\
\hline
\bottomrule
\end{tabular}
\label{table:participant_demographic}
\bigskip
\end{table}

\textbf{Stimuli and material.} This experiment used 5 tests including two subsets of the TIM test, two established face-matching tests (GFMT \cite{burton2010glasgow};
Black-box test,\cite{phillips2018face}), and a standard face-memory test (Cambridge Face Memory Test [CFMT], long form, \cite{duchaine2006cambridge}, \cite{russell2009super}). 
We sampled two 75-item subsets from the original 225-item TIM test to reduce participant fatigue. %Student results from Experiment 1 suggested that ability estimates stabilized at about 75 items \ref{fig:test_reliability}. 
To ensure equal difficulty across the two subsets, we employed item-difficulty measures obtained from IRT modelling in Experiment 1. To do this, we ranked the 225 TIM Test items from least to most difficult and excluded the 4 least difficult items. Next, using the 221 remaining TIM test items, we sampled 75 items randomly, without replacement for each subset.

The Black-box test \cite{phillips2018face} was chosen, because it has been  tested previously on individuals with a wide range of abilities. The test  consists of 20 highly challenging face-matching items. Each item displays two face images of the same identity ($n = 12$) or different identities ($n = 8$). All items displayed frontal-view face images. The task is to determine if the image pairs display the same or different identities using a 7-point scale (+3: Sure they are the same, -3: Sure they are different).  

The GFMT \cite{burton2010glasgow} was selected, because it is considered a common benchmark for face-matching ability.
It includes 40 face-matching items (20 same-identity images pairs and 20 different-identity image pairs. The task is to determine if the image pairs display the same or different identities using binary response options (same or different). 

The CFMT \cite{duchaine2006cambridge} was selected, because it is considered a common benchmark for measuring the ability to identify faces based on memory. It consists of a ``learning phase'' and a ``testing phase''. The learning phase requires the participants to inspect six unfamiliar target faces carefully for memorization. Target faces are presented in one of two possible ways: a) all six face-images in one array or b) separately and consecutively. The testing phase consists of a three-alternative forced choice recognition task, whereby the task is to select the target face  among two other novel faces. The CFMT long form includes 72 items from the original CFMT distributed into three testing blocks (see \cite{duchaine2006cambridge}) and an additional block including 30 very difficult items \cite{russell2009super}). The fourth block in this test was designed to detect higher levels face-recognition ability (e.g., super-recognizers).

\textbf{Procedure.}
Participants completed a total of five tests across two sessions. The tests were divided into two sets: Set A and Set B. Set A included TIM Subset 1, the Black-box test, and the GFMT. Set B included TIM Subset 2 and the CFMT. Sets were counterbalanced such that half the participants completed Set A in session 1 (and Set B in session 2) and half completed Set B in session 1 (and Set A in session 2) \footnote{One participant took the set out of order.}.

During the first session, participants reviewed the consent form with a NIST researcher. The participant was then  assigned randomly to either Set A or Set B. At the end of the first session, the participant completed a demographic questionnaire. Participants returned for the second session approximately one week later (a minimum interval of one week) to complete the second set of tests.

The procedures for the TIM Subset 1 and Subset 2 were the same as for the full item set in Experiment 1. For the GFMT, participants viewed the image pairs and were asked to determine if the pair depicted the same person or different people. On each trial, the images were displayed for 30 seconds. Participants were given unlimited time to respond. For the Black-box test, participants viewed image pairs for up to 30 seconds and were asked to rate the similarity on a 7-point scale (+3: Sure they are the same, -3: Sure they are different). Participants were given unlimited time to respond. For CFMT, participants memorized images for 3 or 20 seconds. Then, they were presented with three images and asked to identify the face that they had seen before. 

\subsection{Experiment 3a: Generalizability across participant population}

{\it Human Performance.} Participant performance was evaluated using two 75-item TIM subsets. We demonstrate that the two TIM subsets occupy a large range of human and item accuracy and that human performance was generalizable across subsets.

Participant accuracy was measured as the proportion of items answered correctly and was above chance (.33) for Subset 1 (\textit{M}= .67, \textit{SD}= .11, \textit{Mdn}= .67) and Subset 2(\textit{M}= .68, \textit{SD}= .11, \textit{Mdn}= .69). Accuracy ranged between 0.41 and 0.91 and between 0.41 and 0.89, for Subset 1 and 2, respectively. We compared participant accuracy on TIM Subset 1 and Subset 2 using a paired sample $t$-test. Results indicated no significant difference (\textit{t}(55)= -0.59, \textit{p} = 0.56, 95\% CI:[-0.03, 0.02]).

Item accuracy was measured as the proportion of participants who answered a given item correctly. For TIM Subset 1, accuracy was above chance (\textit{M}= 0.67, \textit{SD}= 0.16, \textit{Mdn}= 0.68) and ranged between 0.23 and 0.96.
For TIM Subset 2, accuracy was above chance (\textit{M}= 0.68, \textit{SD}=  0.18, \textit{Mdn}= 0.71) and ranged between 0.25 and 0.98.  We compared item accuracy on TIM Subset 1 and Subset 2 using an independent sample \textit{t}-test. Results indicated no significant difference (\textit{t}(145.81)= -0.25, \textit{p} = 0.8, 95\% CI:[ -0.06, 0.05]).

{\it Model.}
We applied IRT modelling and show that the test captures a large range of participant ability and item difficulty. More important, we demonstrate that a model trained on university students and a model trained on non-university students provide comparable ability estimates for non-student individuals. This suggest that non-student participants can be evaluated using smaller sets of items and a model trained on larger data derived from university students.

We trained a one-parameter logistic model (NIST-Model) to evaluate the psychometric properties of the 150-item TIM test using data from NIST employees. The TIM Subsets 1 and 2 were combined into one set of 150 items.  A one-parameter logistic model was fit to the data from the 56 NIST participants and 150 items, hereinafter referred to as the NIST-Model. Results indicated a good fit for the model (RMSEA = 0 , AIC = 9462.686 BIC = 9768.514). Participant ability ranged between -0.97 and 1.14 and item-difficulty ranged between -4.13 and 1.26. 

\begin{figure}[ht!]
\begin{center}
\includegraphics[width=\textwidth]{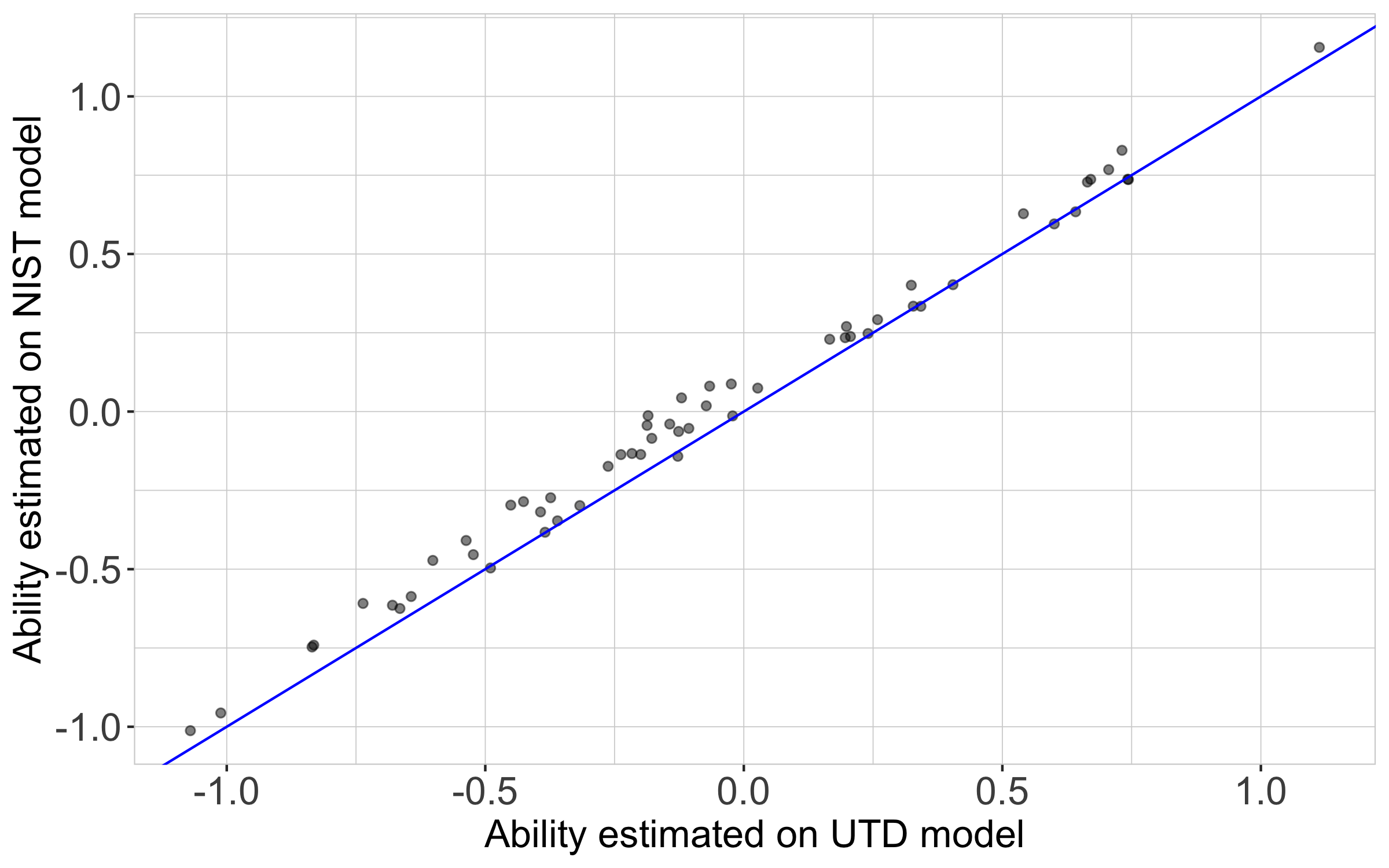}
\caption{Ability scores for NIST participants estimated by the NIST-model (Y-axis) and estimated by the UTD-model (X-axis). Each point represents a NIST participant. Black dots located above the identity line (blue) of the plot indicates that the UTD-model slightly underestimates participant ability in comparison to the NIST-model.}
\label{fig:NISTsubjectAbility_UTDmodelvsNISTmodel}
\end{center}
\end{figure}

We examined whether a one-parameter logistic model trained on university students can be used to estimate participant ability for a separate sample of participants (NIST employees). To do this, we treated the TIM Subsets 1 and 2 as a single 150-item set. Ability scores for NIST participants were estimated using two models. The first set of ability scores was estimated using the NIST-model trained in Experiment 3. The second set of ability scores was estimated using the UTD-trained model from  Experiment 1. Specifically, we projected the responses of all NIST participants (56 participants, 150 items) onto the UTD-Model trained on 197 university students and 225 items. A Pearson's product-moment correlation was used to compare the two sets of ability estimates computed for the NIST participants. Results indicate a strong significant correlation (\emph{r}(54)=.99,  \emph{p} $<$ .001, 95\% CI [0.9999, 0.9999]). Figure \ref{fig:NISTsubjectAbility_UTDmodelvsNISTmodel} illustrates the ability estimated by the NIST-model against the ability estimate by the UTD-model. It is important to note that the data points fall above the identity line, indicating that the ability estimates are slightly underestimated by the UTD-model in comparison to the NIST-model. This result is expected given that the data points illustrated pertain to the same sample of participants used to trained the NIST-model. Overall, these results suggest that a model trained on university student data can generalize to participants from a different population (federal employees), who have been tested in a different experimental setting (NIST).

\begin{figure}[ht!]
\begin{center}
\includegraphics[width=\textwidth]{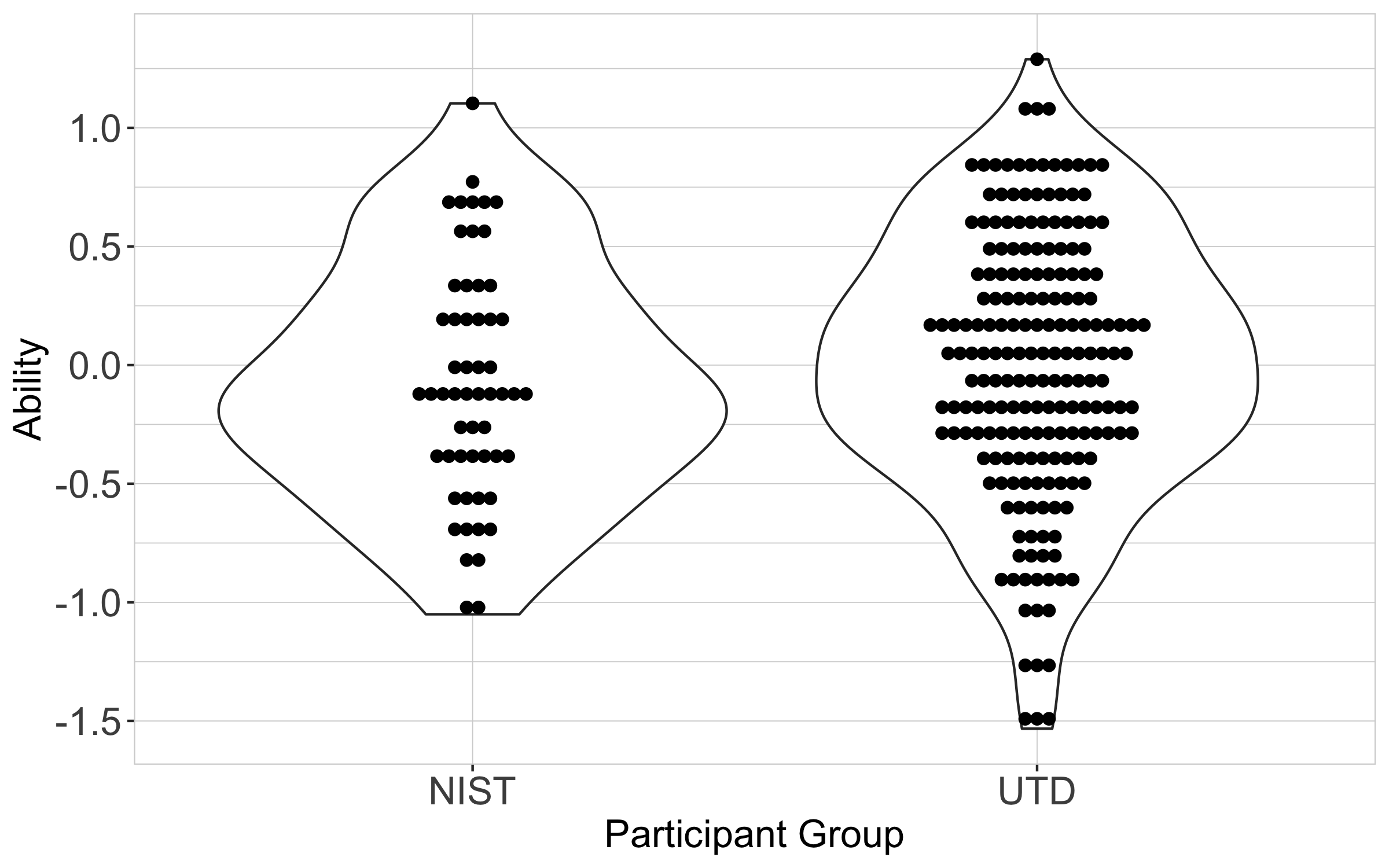}
\caption{Ability scores estimated by the UTD-model for NIST participants (left) and UTD participants (right). Each dot represents a participant.}
\label{fig:groups_UTDmodel}
\end{center}
\end{figure}

\textit{Group comparisons.} Next, we examined whether the two groups of participants produced similar ability measures. All ability scores were estimated using the UTD-model trained in Experiment 1. Specifically, NIST ability scores were estimated by projecting the responses of all NIST participants (56 participants, 150 items) to the UTD-model. All UTD participant ability scores were obtained from Experiment 1. Participant-ability estimates were compared using a Wilcoxon rank sum test. Results indicate no significant difference (\textit{W} = 4976,  \textit{p} = 0.2642). Figure \ref{fig:groups_UTDmodel} illustrates the ability-scores estimated by the UTD-model for each group of participants using the 150-item set. Overall, these results indicate that the two sets of participants do not differ in terms of face-matching ability.

\subsection{Experiment 3b: Generalizability across test session}

Generalizability of the TIM test can be measured, also, by its ability to consistently yield similar results across days. In practice, proficiency subsets should yield similar results at different time points (when no training is involved). To examine the natural variability of individual performance across time, NIST participants completed two equally-difficult tests across two separate testing sessions. We also examine the variability in individual performance across the two equally-difficult tests using UTD student data from Experiment 1. We demonstrated that individual performance varies less (naturally, in the absence of training) across testing sessions than across tests.

Specifically, we examined NIST and UTD individuals performance on two 75-item sets, henceforth referred to as Subsets 1 and 2. As noted, NIST participants completed Subsets 1 and 2 in different testing sessions separated by one week. The order in which the item subsets (1 and 2) were administered was counterbalanced over test sessions. UTD participants completed the full TIM test in a single session (Experiment 1). Figure 3 shows ability estimates derived from Subset 1 and 2. We conceptualized the problem as the follows: The performance of NIST participants across subsets and across testings sessions, was used to estimate variance over a change in session and test ($\sigma^2 _{\Delta S \Delta T}$). Similarly, the performance of UTD participants across subsets on the same session was used to estimate variance over a change in test ($\sigma^2 _{\Delta T}$).  Finally, variance across time ($\sigma^2 _{\Delta S}$)  was solved as described in Equation \ref{time variance model simp.}

\begin{equation}\label{time variance model simp.}
\sigma^2 _{\Delta S}  = \sigma^2 _{\Delta S \Delta T}  - \sigma^2 _{\Delta T} 
 \end{equation}

We estimated variance over a change in session and test ($\sigma^2 _{\Delta S \Delta T}$) using the data from NIST participants and the UTD-model trained in Experiment 1. Specifically, we produced a set of ability scores associated with each Subsets 1 and 2, separately, by projecting NIST-participant responses to each item onto the model. Next, we computed the variance over a change in session and test as the variability in the difference between participant-ability estimates derived from Subsets 1 and 2. Variance over a change in session and test resulted in a value of 0.40. We estimated variance over a change in test ($\sigma^2 _{\Delta T}$) using the data from UTD participants. To do this, we followed the same steps used to compute $\sigma^2 _{\Delta S \Delta T}$ \footnote{Note that ability estimates for UTD participants were computed by projecting a set of responses that were used to train the same model (UTD-model, Experiment 1)}. Results indicated that variance over a change in test is equal to 0.31. Finally, we estimated variance over a change in session using Equation \ref{time variance model simp.}. Variance over a change in session resulted in a value of 0.25. Overall, the results suggest that human participants vary moderately, and that they vary more across tests than across testing session.

\begin{figure}[ht!]
\begin{center}
\includegraphics[width=\textwidth]{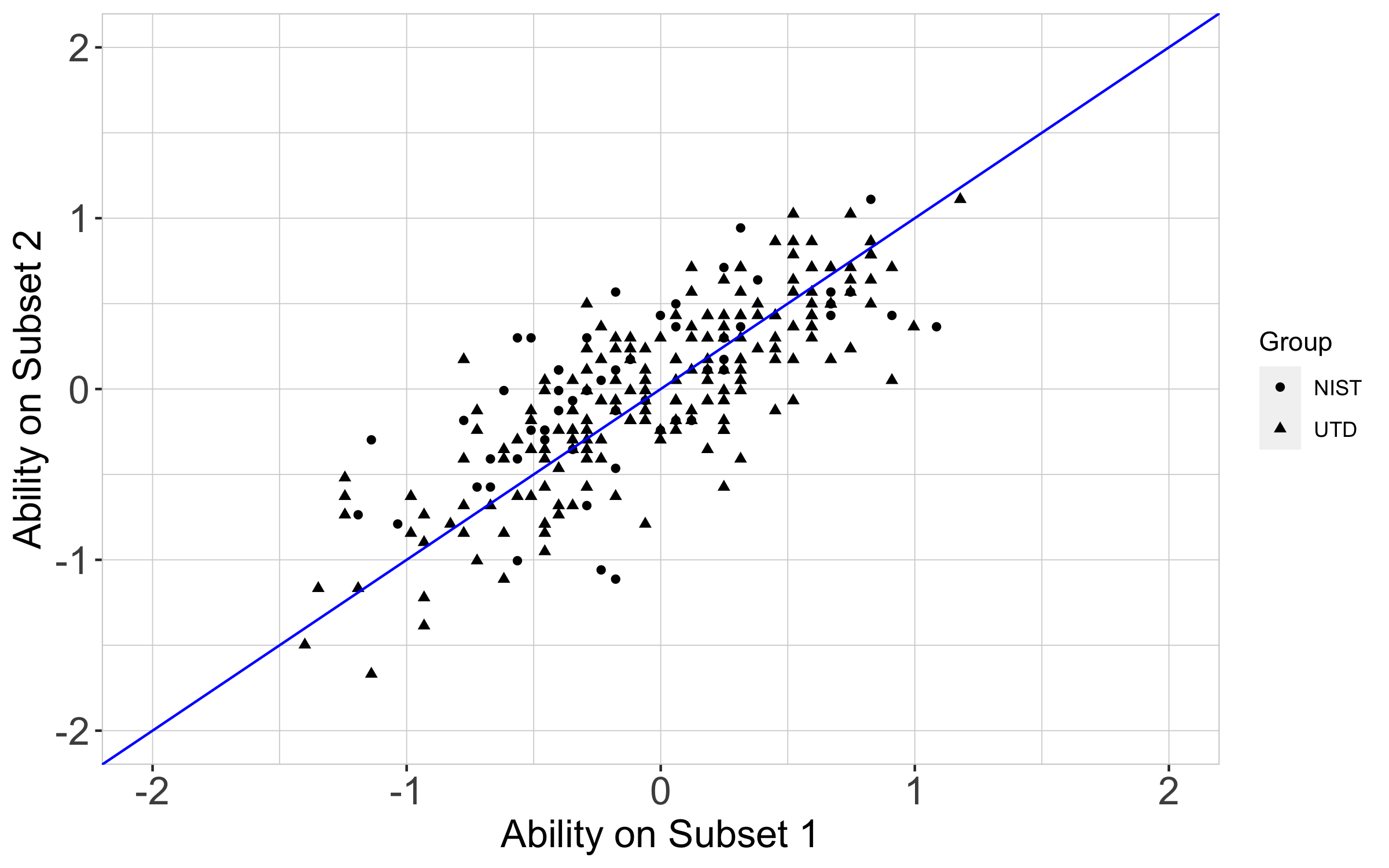}
\caption{Ability scores for NIST participants (dots) and UTD participants (Triangle) derived from Subset 1 plotted against ability scores derived from  Subset 2. All ability measures were estimated using the UTD-model trained in Experiment 1. }
\label{fig:subsets_groups_UTDmodel}
\end{center}
\end{figure}

\subsection{Experiment 3c: Comparability in performance across common face-recognition tests}

In this section, we demonstrate that face-matching ability estimated from a TIM subset (150-item subset) can serve as an indicator of performance for common face-matching (GFMT and Black Box) and face-memory (CFMT) tests. Additionally, we show that the relationship between the TIM test and the CFMT falls on the higher end of the range of correlation coefficient found for other face-matching tests \cite{fysh2018individual, balsdon2018improving, robertson2020super, verhallen2017general, mccaffery2018individual, bobak2016super, wilmer2012capturing}

First, the TIM Subsets 1 and 2 were combined into one set of 150 items. We computed Pearson product-moment correlations to examine the relationship between individual performance across all tests. We measured individual accuracy on the GFMT and CFMT tests as proportion correct. We measured individual accuracy on the Black Box test as the AUC. To estimate individual ability from the 150-item set (TIM test), we projected the responses of all NIST participants onto the UTD-model trained in Experiment 1. Pearson's product-moment correlation results indicated a significant and moderate relationship between the 150-item TIM test and the face-matching tests [GFMT: \emph{r}(54)= 0.45,  \emph{p} $<$ .001, 95\% CI [ 0.21, 0.64], Black Box: \emph{r}(54)= 0.45,  \emph{p} $<$ .001, 95\% CI [ 0.22, 0.64]. Also, Pearson's product-moment correlation results indicated a significant and moderate relationship between the 150-item TIM test and the CFMT (\emph{r}(54)= 0.59, \emph{p} $<$ .001, 95\% CI [0.39, 0.74]). Moreover, results indicated a moderate correlation between the Black Box and the GFMT (\emph{r}(54)= 0.42, \emph{p} = .001, 95\% CI [0.18, 0.61]) and a weak correlation between the Black Box and the CFMT (\emph{r}(54)= 0.38, \emph{p} = .003, 95\% CI [0.14, 0.59]). Lastly, results indicated a moderate correlation between the CFMT and the GFMT (\emph{r}(54)= 0.57, \emph{p} $<$ .001, 95\% CI [0.36, 0.72]). The findings indicate that the relationship between the TIM test and other tests falls within the same range of correlation coefficients found in previous work \cite{fysh2018individual, balsdon2018improving, robertson2020super, verhallen2017general, mccaffery2018individual, bobak2016super, wilmer2012capturing}. Overall, these findings suggest that ability estimates derived from the TIM test can indicate performance on more common tasks such as face matching (e.g., GFMT, Black Box) and memory-based face recognition (e.g., CFMT).

\subsection{Experiment 3 Discussion}

The goal of Experiment 3 was to evaluate the generalizability of the TIM test across participant groups and across testing time, and to evaluate its comparability to commonly-used face recognition assessment tools. Overall, our findings indicate that the psychometric properties of the test remain consistent across a different group of participants (university students from UTD and federal employees from NIST) and a different testing setting (UTD laboratory and NIST laboratory). We also demonstrated that an IRT-model trained on the full TIM test and a large sample of university students can be used to evaluate NIST employees using a smaller item set. This experiment also provides a proof principle of the applicability of the TIM test and IRT for assessing changes in individual ability across time. Finally, we demonstrate that face-matching ability estimated from the TIM test is correlated with performance on commonly used face-matching and face-memory tests. 
 
\section{Section 6: General Discussion and Conclusion}

The objective of this study was to refine the current state of face identification testing by developing a framework for creating proficiency tests. This framework relies on IRT to calibrate item difficulty in relation to participant ability, thereby enabling the selection of subsets of items that can be combined in systematic ways to create tests of specified difficulty. These item subsets can be tailored for testing individuals of specific ability levels and for testing professionals who are busy and may only be able to spare time for short tests. Multiple tests of equal difficulty can be used also to detect {\it changes} in ability (e.g.,from training, experience, or age). Because items are not reused in multiple tests, proficiency improvements can be detected without confounding factors that result from repeated exposure to the same faces.

Using this framework, we introduce the TIM test, which includes items that span a range of difficulty from very easy (97\% of participants endorsed a correct response to the item) to very challenging (17\% of participants endorsed a correct response to the item). This range of difficulty supports the assessment of participant abilities close to random performance (accuracy of 37\%) to high ability (accuracy of 89\%). The TIM test was designed to address longstanding response bias issues in traditional face identification tests due to the use of rating scales and binary decision choices. Response bias poses a particularly vexing problem when comparing across groups of different ability who use the scale in different ways. The TIM test stimuli and materials that support the framework (de-identified data and code to build the student-based one parameter model) can be obtained for research use.\footnote{Images used in the TIM test are available by license from the University of Notre Dame. R code to fit the one-parameter logistic model and run the analysis, de-identified participant responses, as well as PsychoPy experimental code will be made available by the UTD research team at OSF website.} The framework and results we present provide a general foundation for future research that connects to basic theory in the psychology of face recognition, as well as to testing in research and applied scenarios.

It has become increasingly clear in the psychological literature that successful face identification requires two important skills. The first is the need to discriminate highly similar faces  (i.e., ``telling people apart'')---long considered the basis for human expertise with faces \cite{diamond1986faces}. The second, is the ability to perceive identity constantly across multiple face images that vary in appearance and image conditions (e.g., expression, viewpoint, illumination)(i.e., ``telling faces together'') \cite{jenkins2011variability,andrews2015telling}. The constructed triads used in the TIM test implicitly test both skills simultaneously. In particular, the triads  evaluate both the ability to ``tell people together'' and ``tell people apart'' with stimuli constructed to be challenging for both tasks.

The framework developed in this study, combined with the TIM test we introduce, provides a path for building calibrated face-identification tests. 
To establish a general baseline and proof of concept, the current study was limited to university students and federal employees. Although simulation results (See supplemental) offer {\it prima facie} evidence that the test transferred well between students of higher and lower ability, this should be verified explicitly with other populations. Future research should focus on evaluating the TIM test for non-student populations such as forensic examiners, super-recognizers, forensic specialists, and prosopagnosics.

One motivation for developing this framework concerned the challenges of measuring item difficulty in test paradigms such as identity matching. These paradigms allow for user response bias in the form of rating scales or binary choice decision options and are some of the most commonly used in forensic practice. A requirement of an efficient face identification proficiency test is to provide measures of ability that can translate to people's proficiency for applied settings. Therefore, it is important to verify that measures of proficiency gleaned from a 3AFC test, such as the TIM test, accurately predict performance in identity-matching tasks.

Finally, expecting that the TIM test can spur extensive research in the face-identification community, we make the test available online to researchers. Other researchers can test individuals using the full set of items or the subsets of items used here to estimate the abilities of individuals. This supports easy comparison with the student data we report here. Using the existing model, researchers can project their data to estimate ability for other populations of interest, or they can merge their data to create a new model.

In anticipating the future of calibrated face identification tests, future research should build on the approach proposed here and examine more complex IRT models (e.g., two- and three-parameter model \cite{birnbaum1968some}). We based our study on the one-parameter logistic model, which do not model participant guessing and assumes all items have equal discrimination parameters. More general IRT models were developed to handle both conditions. These models would offer a deeper understanding on participants’ ability and item difficulty and contribute to designing well-calibrated proficiency tests. Understanding the nature of participant ability and item difficulty would offer a starting point for developing adaptive face identification tests (e.g., Computerized Adaptive Tests).

Although this study focused on face-identification, nothing in our framework is specific to facial comparisons. Researchers and practitioners can apply our work to disciplines that perform comparisons, for example, latent fingerprint, speaker, and iris identification. Our method has the potential to provide multiple forensic disciplines with the tools to create calibrate proficiency tests.

\section{TIM test availability}
The TIM test will be made available for research purposes, without cost, by license from the University of Notre Dame. Specifically, researchers will be able to access all TIM test images from a repository, provided by the University Notre Dame, after signing a license in which they agree to the conditions of use. This process ensures that all individuals who wish to use the test accept responsibility for adhering to subject protections and protecting subject privacy. 

\begin{verbatim}
https://cvrl.nd.edu/projects/data/#triad-identity-matching-tim-test-data-set
\end{verbatim}
Other materials (R code to run the analysis, de-identified data, and PsychoPy experimental code) will be made available by the UT Dallas research team at OSF website.

\bibliographystyle{unsrt}  
\bibliography{references}

\end{document}